%% file: piveted_granite.tex
\documentclass[sigplan]{acmart}
\usepackage{booktabs} % For formal tabl
\usepackage{caption}
\usepackage{subcaption}

\graphicspath{{figures/}}
\setcopyright{rightsretained}
\usepackage{multirow}
\newcount\Comments   % 0 suppresses notes to selves in text
\usepackage{color}
\definecolor{purple}{rgb}{1,0,1}
\newcommand{\kibitz}[2]{\ifnum\Comments=1\textcolor{#1}{#2}\fi}

\newcommand{\nth}[1]{\ensuremath{#1^{\text{th}}}}
\usepackage[ruled]{algorithm2e}
\usepackage{tensor}

\newcommand{\methodName}{PIVETed-Granite}

\usepackage{bbm}

\acmConference[MLMH'18]{2018 KDD Workshop on Machine Learning for Medicine and Healthcare}{August 2018}{London, United Kingdom}
\begin{document}
\title[\methodName]{\methodName: Computational Phenotypes through Constrained Tensor Factorization}%: A Case Study using Computational Phenotyping}
%\titlenote{Produces the permission block, and
 % copyright information}
%\subtitle{Extended Abstract}
%\subtitlenote{The full version of the author's guide is available as
%  \texttt{acmart.pdf} document}

\author{Jette Henderson}
\authornote{jette@ices.utexas.edu}
\affiliation{
  \institution{University of Texas, Austin}}
%  \streetaddress{P.O. Box 1212}
%\email{jette@ices.utexas.edu}
\author{Bradley A. Malin}
\affiliation{
  \institution{Vanderbilt University}}

\author{Joyce C. Ho}
\affiliation{
  \institution{Emory University}}
 % \streetaddress{P.O. Box 1212}
 %% \postcode{43017-6221}
%}
%\email{webmaster@marysville-ohio.com}

\author{Joydeep Ghosh}
%\authornote{This author is the
%  one who did all the really hard work.}
\affiliation{%
  \institution{University of Texas, Austin}}
%  \streetaddress{1 Th{\o}rv{\"a}ld Circle}
%\email{larst@affiliation.org}

\renewcommand{\shortauthors}{J. Henderson et al.}

\begin{abstract}
%Multiway data can be succinctly contained in multiway arrays.
It has been recently shown that sparse, nonnegative tensor factorization of multi-modal electronic health record data
is a promising approach to
high-throughput computational phenotyping.
However, such approaches typically do not leverage available domain knowledge while extracting the phenotypes; hence,
some of the suggested phenotypes may not map well to clinical concepts or may be very similar to other suggested phenotypes.
To address these issues, we present a novel, automatic approach called
 \methodName~that mines existing biomedical literature (PubMed) to obtain cannot-link constraints that are then used as side-information during a tensor-factorization based computational phenotyping process.
The resulting improvements are clearly observed in experiments using a large dataset from VUMC to identify phenotypes for hypertensive patients.

\end{abstract}


%\ccsdesc[500]{Computing methodologies~Machine learning~Machine learning approaches~Factorization methods}
%ccsdesc[300]{Computing methodologies~Machine learning~Machine learning algorithms~Regularization}


%  \includegraphics[width=\textwidth]{ami_phenos_trim_long.png}
%  \caption{\methodName phenotypes derived from a tensor constructed from VUMC patient-level data. These phenotypes have high membership of patients who had at least one myocardial infarction.}
%  \label{fig:ami_phenos}

\maketitle

\input{sections/introduction_v3}
\input{sections/methods_v3}

\input{sections/results_v2}

\input{sections/discussion_v2}

\bibliographystyle{ACM-Reference-Format}
\vspace{-.1in}
\bibliography{sections/cp_clic}

\end{document}

%% file: sections/introduction_v3.tex
\vspace{-.1in}
\section{Introduction}
%\begin{figure}[h!]
%\includegraphics[width=\linewidth]{ami_phenos_trim}
%\caption{\methodName~-derived phenotypes with high percentages of patients who at least one myocardial infarction.}
%\label{fig:ami_phenos}
%\end{figure}
%\vspace{-.05in}
\begin{figure*}
  \includegraphics[width=\textwidth]{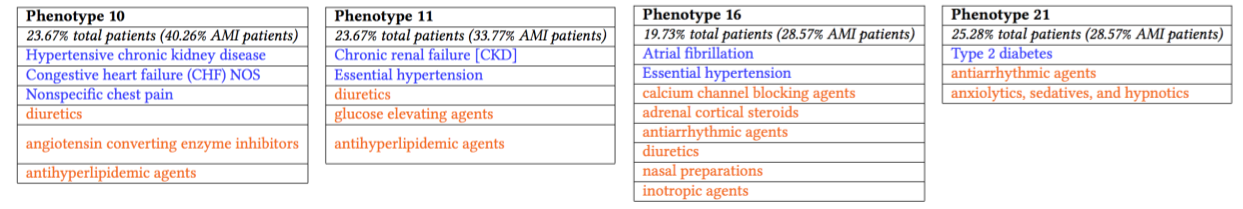}
  \caption{\methodName phenotypes derived from a tensor constructed from VUMC patient-level data. These phenotypes have high membership of patients who had at least one myocardial infarction.}
  \label{fig:ami_phenos}
  \end{figure*}

Computational phenotyping is the process of algorithmically deriving cohesive sets of clinical characteristics from collections of patient documentation like electronic health records (EHRs)~\cite{pathak2013electronic}.
For a set of computational phenotypes to be useful to clinicians, they should be 1) sparse (i.e., have relatively few elements) and 2) diverse (i.e., have few overlapping elements).
Most importantly, they should map to clinically relevant concepts.
Variations of a tensor factorization method called CANDECOMP/PARAFAC (CP) decomposition have shown potential in deriving computational phenotypes with these characteristics.
CP decomposition factors multiway arrays, or tensors, into sets of rank-one components.
In the application of computational phenotyping, the non-zero elements of each component can be interpreted as elements of a phenotype.
Figure~\ref{fig:ami_phenos} shows examples of computational phenotypes derived through tensor factorization.
\citet{Ho:2014da} were the first to show tensor decomposition could be applied to tensors constructed from count data extracted from EHRs to derive phenotypes, a large number of which were clinically relevant.
Subsequent models have been developed with the goal of deriving sparse, diverse, and interpretable phenotypes~\cite{Ho:2014jc,wang2015rubik,henderson2017c}.
\citet{wang2015rubik} incorporated guidance information from domain experts into the tensor decomposition process, but obtaining input from domain experts may not always be possible.
\citet{henderson2017c} introduced a CP model called Granite with angular constraints to encourage diverse phenotypes and sparsity constraints to derive succinct phenotypes but found there was a trade-off between the diversity and the clinical meaningfulness of components.

One possible weakness of using CP decomposition is that there can be noise between and across the modes (i.e., elements appear together that do not belong together).
In computational phenotyping, this noise could manifest as a medication and diagnosis co-occurring in a component but not actually having a clinical relationship. 
This weakness can degrade the interpretability of the fits~\cite{henderson2017c}.
Few tensor decomposition methods applied to clinical data have used supervision or domain expertise to increase the number of interpretable components~\cite{wang2015rubik}.
~Like many problems in machine learning, incorporating supervision can be challenging and costly in terms of the time and domain expertise necessary for gathering labels or domain-specific constraints.

In this work, we explore a new proxy for domain-expertise via side information extracted using PIVET~\cite{henderson2018a}, a phenotype validation tool.
The goal is to increase the number of meaningful components in the CP decomposition process without human input.
PIVET analyzes a corpus of publicly available medical journal articles to create evidence sets for candidate phenotypes.
Our method, which we refer to as \methodName,~ involves a novel application of PIVET to provide side information in the form of a cannot-link matrix between the diagnosis and medication modes of a tensor constructed from a set of EHRs.
By automatically leveraging available biomedical literature, \methodName~enables more concise and diverse phenotypes that are discriminative and interpretable.

%% file: sections/methods_v3.tex
\vspace{-.1in}
\section{\methodName~Framework}
\emph{Background.}
We briefly describe tensors and the process of decomposing them but refer the reader to~\cite{Kolda:2009dh} for a more detailed account.
A tensor is an $n$-way or $n$-mode array that is used to represent $n$-dimensional relationships.
We use CANDECOMP/PARAFAC (CP) decomposition \cite{Harshman:1970vk} to factor tensors.
For a tensor $\TX{X}$ with $N$ modes, the CP decomposition is
\begin{align}
\TX{X} &\approx \displaystyle \sum_{r=1}^R \lambda_r \M{a}_r^{(1)} \op \hdots \op \M{a}_r^{(N)} = \llbracket \BS{\lambda}; \FN{A}{1}; \hdots; \FN{A}{N} \rrbracket,
\end{align}
where each component of the sum is a rank-one tensor formed by taking the outer product of N vectors, $\FN{a}{1} \op \FN{a}{2} \op \cdots \op \FN{a}{N}$.
The representation $\llbracket \BS{\lambda}; \FN{A}{1}; \hdots; \FN{A}{N} \rrbracket$ is shorthand notation with the weight vector $\BS{\lambda} = [\lambda_1 \cdots \lambda_R]$ and the factor matrix $\M{A}^{(n)} = [\FN{a}{n}_1 \cdots \FN{a}{n}_R],$ where $\M{a}_r$  denotes the \nth{r} column of $\M{A}^{(n)}$.
CP decompositions are usually fit using a loss function that makes assumptions about the underlying distribution that generated the data in the tensor. %(e.g., Gaussian or Poisson).
When CP decomposition is applied to tensors constructed from EHRs, each rank-one tensor can be thought of a phenotype.

\emph{Problem Formulation.} \methodName~is built on existing nonnegative CP decomposition algorithms that model the observed data using the Poisson distribution\cite{Ho:2014jc,Ho:2014da,henderson2017c}.
We focus on a $3$-mode tensor where the three dimensions are (1) patients, (2) diagnoses, and (3) medications, and each element is a count of the number of times a patient received a diagnosis and medication prescription in a given period of time.
An observed tensor, $\TX{X} \in \mathbb{R}^{I_1 \times I_2 \times I_3}$ is approximated as the sum of $R$ $3$-way rank-one tensors $\TX{X} \approx \TX{Z} = \llbracket \BS{\lambda}; \M{A}; \M{B} ; \M{C} \rrbracket$, which are the patient, diagnosis, and medication factor matrices, respectively.
To discourage specified diagnosis and medication pairs from appearing together in the same phenotype, \methodName~introduces a cannot-link matrix between the diagnosis ($\M{B}$) and the medication ($\M{C}$) factor matrices. 
The optimization problem for the observed tensor $\TX{X}$ is:
\begin{align}
f(\TX{X}) &= \min ( \displaystyle \sum_{\V{i}} ({\fv{z}}  -  {\fv{x}} \log {\fv{z}})\label{eq:kl} \\
&~~~~+ \frac{\beta_{1}}{2} \sum_{r=1}^R \sum_{p=1}^r \left((\max\{0,\frac{(\M{d}_{p})^\tr \M{d}_r}{||\M{d}_{p}||_2 ||\M{d}_{r}||_2}-\theta_{\M{d}}\})^2 \right) \label{eq:ang_pen}\\ 
&~~~~+ \frac{\beta_2}{2} \sum_{r=1}^R (||\M{a}_{r}||_2^2 + ||\M{b}_{r}||_2^2 + ||\M{c}_{r}||_2^2) \label{eq:l2}\\ 
&~~~~+ \beta_{3} \text{trace}(\M{B}^{\tr} \M{M} \M{C}) )\label{eq:cl}\\
\text{s.t }  \TX{Z} &= \llbracket \sigma; \M{u}_a; \M{u}_b; \M{u}_c \rrbracket +\llbracket \BS{\lambda}; \M{A}; \M{B}; \M{C} \rrbracket\\
&~~~~\M{d} \in \{\M{a},\M{b},\M{c}\} \\ 
&~~~~||\M{a}_r||_1 = ||\M{b}_r||_1 = ||\M{c}_r||_1 = 1, \M{a}_r, \M{b}_r, \M{c}_r \geq 0 \\
&~~~~||\M{u}_{a}||_1  = ||\M{u}_{b}||_1  = ||\M{u}_{c}||_1 =1,  \M{u}_a, \M{u}_b, \M{u}_c > 0.
\end{align}

For count data, the loss function is KL-divergence (\ref{eq:kl}).
An angular penalty term (\ref{eq:ang_pen}) discourages any factors from being too similar, where similarity is defined as the cosine angle between two factor vectors, and an $l_{2}$ penalty term controls the growth of the size of the factors (\ref{eq:l2}) (See ~\cite{henderson2017c}~for details).

\emph{Incorporating PIVET.} In Equation~\ref{eq:cl},  $\M{M} \in  \M{1}^{I_2 \times I_3}$ is a binary cannot-link matrix defined as follows:
\[ 
\M{M}_{jk}    = \begin{cases}
                                   1, & \text{if lift(${b}_{j},{c}_{k}$) $< \alpha$)} \\
                                   0, & \text{otherwise}  \end{cases}
\]
We construct $\M{M}$ with PIVET.
PIVET calculates the lift for each (diagnosis, medication) pair (i.e., ${b}_{j},{c}_{k}$) based on analysis of biomedical journal articles.
A lift of much greater than $1$ indicates diagnosis $j$ and medication $k$ co-occur often and therefore may have a clinical relationship with one another, and a value of $1$ or less means diagnosis $j$ and medication $k$ do not co-occur often in the corpus and may not have a clinical relationship.
In this work, we use $\alpha = 1$.
The terms in Equation~\ref{eq:cl} are of the form ${b}_{jr} \M{M}_{jk} {c}_{kr}$, and only contribute to the objective function if the $j^{\text{th}}$ diagnosis and the $k^{\text{th}}$ medication appear in the $r^{\text{th}}$ component.
Since (\ref{eq:cl}) is a soft constraint if there is actually a relationship between $({b}_{j},{c}_{k})$ in the data, they can still appear together in components.
However, if the relationship is weak in the data, these elements will be discouraged from appearing together in a phenotype. 

We solve for $\TX{Z}$ using gradient descent.
We refer the reader to~\cite{henderson2017c} for derivation of the gradients for Equations~\ref{eq:kl},~\ref{eq:ang_pen},~\ref{eq:l2}, and~\cite{petersen2008matrix} for the gradient of Equation~\ref{eq:cl}.

%% file: sections/results_v2.tex
\begin{figure}[h!]
\includegraphics[width=.7\linewidth]{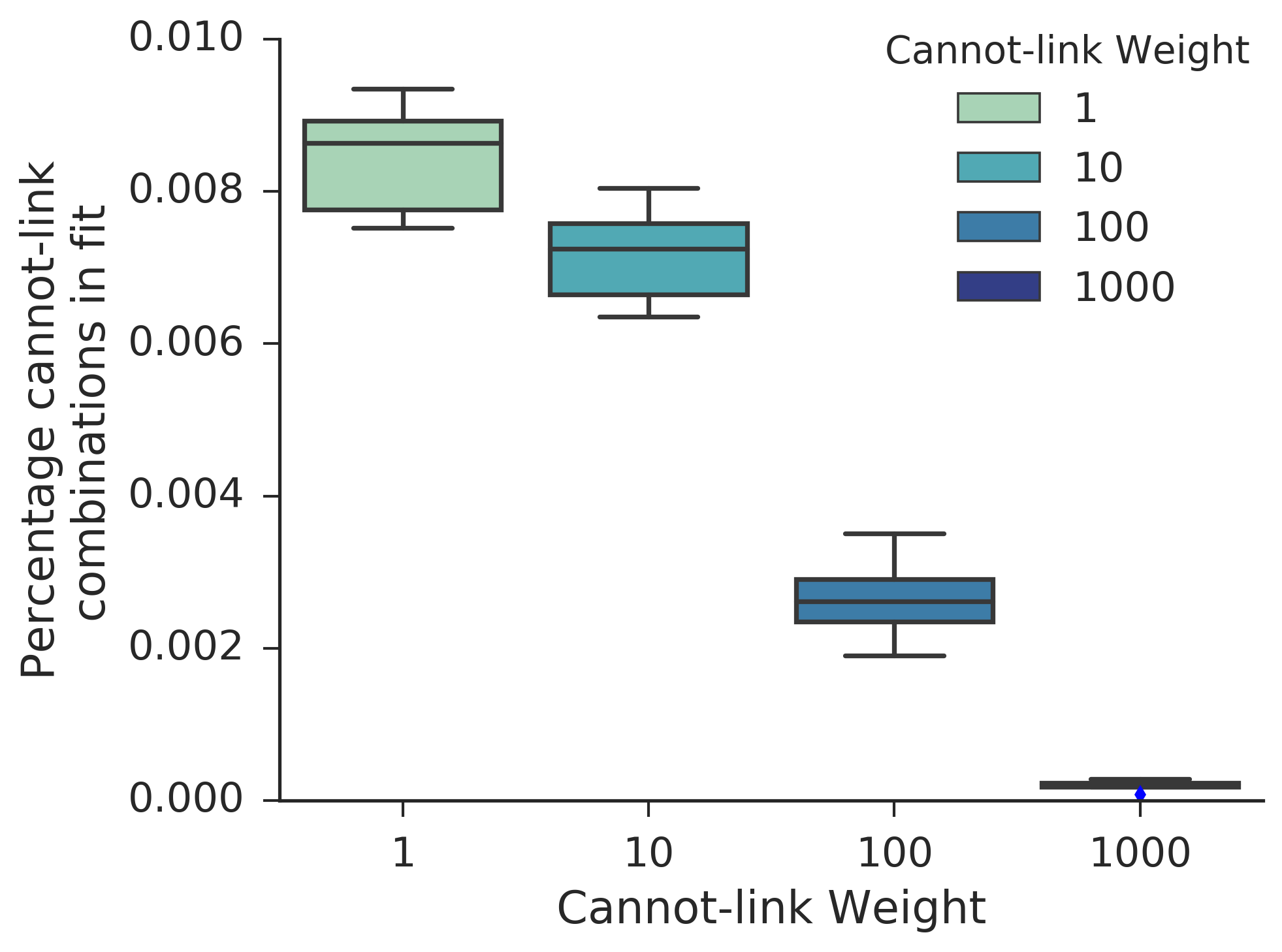}
\caption{Percentage of (diagnosis, medication) cannot-link constraints appearing in the final fit.}
\label{fig:cl_weight}
\end{figure}

  \begin{figure*}
  \includegraphics[width=.65\linewidth]{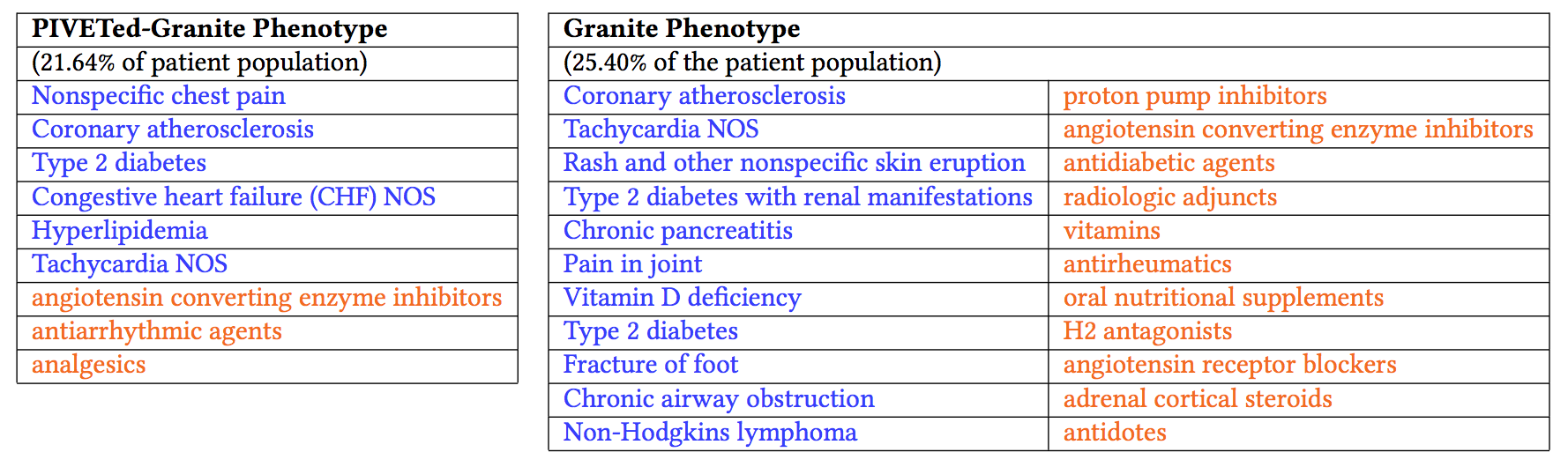}
  \caption{Two phenotypes, one derived using \methodName~(left) and one using Granite (right) where both methods were initialized with the same factor vetors.}
  \label{fig:pheno_comp}
\end{figure*}

\vspace{-.1in}
\section{Experiments}
\label{sec:exps}
\emph{Dataset Description}. To explore the feasibility of using guidance from PIVET, we constructed a tensor from the diagnosis and medication counts of 1622 patients from the Synthetic Derivative (SD), a de-identified EHR database gathered at the VUMC\cite{Roden:08}.
A panel of domain experts developed sets of characteristics (i.e., billing and medical codes) to identify patients as case and control for a set of diseases\cite{ritchie2010robust}. 
Using these specifications, we included 304 resistant hypertension case patients and 399 resistant hypertension control patients in the tensor.
For each case patient, we counted the medication and diagnosis interactions that occurred two years before they received the hypertension diagnosis. 
For each non-case patient, we counted the interactions that occurred two years before their last interaction with the VUMC.
In their raw form, the diagnosis codes (International Classification of Diseases (ICD-9) system) capture a very detailed level of information. 
We use PheWAS coding to aggregate the diagnosis codes into broader categories\cite{denny2013systematic} and Medical Subject Headings (MeSH) pharmacological terms from the RxClass RESTful API to group the medications into more general categories\footnote{https://rxnav.nlm.nih.gov/RxClassAPIs.html}.
These coarser hierarchies produced a tensor with 1622 patients by 1325 diagnoses by 148 medications.

\begin{table*}[]
\centering
\caption{Fit information for phenotypes derived using Marble, Granite, and \methodName.}
\label{tab:fit_stats}
\begin{tabular}{@{}llllllll@{}}
\toprule
                &                        & \multicolumn{3}{c}{\textbf{Average Number of Non-Zeros}}    & \multicolumn{3}{c}{\textbf{Cosine Similarity}}              \\ 
\textbf{Method} & \textbf{KL-divergence} & \textbf{Patient} & \textbf{Diagnosis} & \textbf{Medication} & \textbf{Patient} & \textbf{Diagnosis} & \textbf{Medication} \\\midrule
Marble          & 2803253.42 (194914.35) & 26.72 (1.3)      & 7.01 (0.37)        & 8.44 (0.22)         & 0.07 (0.01)      & 0.01 (0.01)         & 0.24 (0.01)         \\
Granite         & 2311866.35 (27826.92)  & 70.45 (1.69)     & 18.21 (1.06)       & 12.31 (0.31)        & 0.18 (0.01)      & 0.02 (0.01)        & 0.12 (0.01)         \\
\methodName                & 2224824.04 (19758.83)  & 57.21 (2.65)     & 5.78 (0.2)         & 5.89 (0.56)         & 0.20 (0.02)       & 0.03 (0.02)        & 0.05 (0.02)         \\ \bottomrule
\end{tabular}
\end{table*}

\emph{Quantitative Evaluation.}
We compare \methodName~to two baseline models, Granite and Marble~\cite{Ho:2014da}.
Table~\ref{tab:fit_stats} shows fit quality (KL-divergence), sparsity, and diversity measures for the three models ($R=30$).
\methodName~was the best fit to the data with the lowest KL-divergence, and also resulted in the smallest number of non-zero elements in the diagnosis and medication modes.
Additionally, \methodName~resulted in diagnosis factors that were comparably diverse to those of Granite and medication factors that were more diverse than Granite.
We also evaluate the effect of the cannot-link weight $\beta_3$ on the percentage of (diagnosis, medication) cannot-link pairs present in the factor matrices.
In Figure~\ref{fig:cl_weight}, as $\beta_3$ increases, the percentage of cannot-link pairs decreases.

Additionally, we evaluated the discriminative capabilities of \methodName~in a prediction task where the patient factor matrix $\M{A}$ served as the feature matrix.
We compared the performance of \methodName, Granite, and Marble using logistic regression to predict which patients were hypertension case and control.
The model ran with five 80-20 train-test splits, and the optimal LASSO parameter for the model was learned using 10-fold cross-validation.
Table~\ref{tab:pred} shows the AUC for \methodName, Granite, and Marble.
The patient factor matrix derived using \methodName~resulted in the most discriminative model in this task.

\emph{Qualitative Exploration.}
To evaluate the effect of the cannot-link matrix $\M{M}$ on the decomposition process we initialized \methodName~and Granite fits with the same factors and then examined the differences between the fitted factors.
Figure~\ref{fig:pheno_comp} shows one phenotype from each method initialized from the same factors.
While the phenotypes are similar to one another, \methodName's characteristics form a more succinct, focused characterization of heart disease complicated with type 2 diabetes. 
Additionally, the Granite phenotype contains many cannot-link combinations (e.g., (``Fracture of foot", ``antidotes")) whereas the \methodName~phenotype does not.
The cannot-link constraints seem to result in phenotypes that are descriptive and cohesive.

As a way to qualitatively explore the clinical meaningfulness of the discovered phenotypes we identified patients who experienced acute myocardial infarctions (AMI), which resulted in a cohort of $77$ unique patients within the tensor.
In Figure~\ref{fig:ami_phenos}, we show the phenotypes with the highest proportions of AMI patients.
These automatically generated phenotypes seem to give nuanced descriptions of patients who have AMIs.
For example, in Phenotype $10$ one of the diagnoses is congestive heart failure, which is primarily caused by acute myocardial infarctions~\cite{cahill2017heart}.
Type-2 diabetes patients (Phenotype $21$) are also more likely to experience heart attacks and have more negative outcomes from them~\cite{lago2009type}.

\begin{table}
\centering
\caption{AUC for predicting resistant hypertension case patients.}
\vspace{-.1in}
\label{tab:pred}
  \begin{tabular}{ l  c     }
  \toprule
    \textbf{Method}    & \textbf{ AUC (st. dev.)}\\
     \midrule
      Marble &  .6656 (.09) \\ 
Granite &   .7083  (.04)\\
\methodName &    {\bf .7172 (.01)}\\ 
\bottomrule
  \end{tabular}
\end{table}

%% file: sections/discussion_v2.tex
\vspace{-.15in}
\section{Discussion and Conclusion}
Adding guidance in the form of constraints to computational phenotyping models can help improve the quality of the fit and shows promise in increasing the clinical meaningfulness of derived phenotypes.
However, obtaining informative constraints can be difficult and expensive in regard to time and effort required by domain experts.
We show how to leverage publicly available information in the form of medical journals to guide the decomposition process to discriminative and interpretable phenotypes.
\methodName~derived phenotypes that were more discriminative, more diverse, and sparser than two competing baseline models.
In the future, we plan to further analyze the clinical interpretability of the \methodName-derived phenotypes and experiment with the threshold used to create the cannot-link constraint matrix.
Incorporating cannot-link constraints between modes is a general method that can be applied to many domains.
In this application, the quality of the auxiliary information provided by PIVET seems to be high, but in other applications, it may not be.
Our next step is to study how to incorporate auxiliary information when that side information is noisy.